\documentclass{article}
\usepackage{ijcai26}

\usepackage{times}
\usepackage{soul}
\usepackage{url}
\usepackage[hidelinks]{hyperref}
\usepackage[utf8]{inputenc}
\usepackage[small]{caption}
\usepackage{graphicx}
\usepackage{amsmath}
\usepackage{amsthm}
\usepackage{booktabs}
\usepackage{algorithm}
\usepackage{algorithmic}
\usepackage[switch]{lineno}

\usepackage{amsfonts}
\usepackage{booktabs}
\usepackage{bm}
\usepackage{subcaption}     

\newtheorem{theorem}{Theorem}

\title{MiCA: A Mobility-Informed Causal Adapter for Lightweight Epidemic Forecasting}

\author{
    Author Name
    \affiliations
    Affiliation
    \emails
    email@example.com
}

\author{
Suhan Guo$^{1,2}$
\and
Jiahong Deng$^{1,2}$
\and
Furao Shen$^{*1,2}$
\affiliations
$^1$State Key Laboratory for Novel Software Technology, Nanjing University, China\\
$^2$School of Artificial Intelligence, Nanjing University, China\\
\emails
\{shguo, jiahongdeng\}@smail.nju.edu.cn, frshen@nju.edu.cn
}

\begin{document}

\maketitle

\begin{abstract}
    Accurate forecasting of infectious disease dynamics is critical for public health planning and intervention. Human mobility plays a central role in shaping the spatial spread of epidemics, but mobility data are noisy, indirect, and difficult to integrate reliably with disease records. Meanwhile, epidemic case time series are typically short and reported at coarse temporal resolution. These conditions limit the effectiveness of parameter-heavy mobility-aware forecasters that rely on clean and abundant data. In this work, we propose the Mobility-Informed Causal Adapter (MiCA), a lightweight and architecture-agnostic module for epidemic forecasting. MiCA infers mobility relations through causal discovery and integrates them into temporal forecasting models via gated residual mixing. This design allows lightweight forecasters to selectively exploit mobility-derived spatial structure while remaining robust under noisy and data-limited conditions, without introducing heavy relational components such as graph neural networks or full attention. Extensive experiments on four real-world epidemic datasets, including COVID-19 incidence, COVID-19 mortality, influenza, and dengue, show that MiCA consistently improves lightweight temporal backbones, achieving an average relative error reduction of 7.5\% across forecasting horizons. Moreover, MiCA attains performance competitive with SOTA spatio-temporal models while remaining lightweight. Code is available at: \url{https://anonymous.4open.science/r/MiCA-DF48}
\end{abstract}

\section{Introduction}
Accurate forecasting of daily case numbers for infectious diseases is crucial for disease management. The exponential growth of cases serves as an indicator of upcoming waves and the emergence of new variants~\cite{elliottExponentialGrowthHigh2021}. 
Such forecasts allow governments to update and implement directives, facilitating proactive resource allocation to prevent strain on healthcare systems.

Human mobility is a major driver of spatial dependence in infectious disease dynamics. 
Population movement across regions induces strong cross-regional correlations in reported case trajectories, making mobility information highly relevant for forecasting tasks~\cite{colizzaRoleAirlineTransportation2006,jiaPopulationFlowDrives2020}. In practice, however, mobility is inherently difficult to measure and often contaminated by noise. 
Meanwhile, infectious disease surveillance data are costly to collect and maintain, and are frequently reported at low temporal resolution with missing or erroneous values. 
Under these conditions, training expressive mobility-aware forecasters that rely on clean and abundant data becomes impractical, motivating the need for alternative modeling strategies.

\begin{figure}[t]
    \centering
    \includegraphics[width=0.5\textwidth]{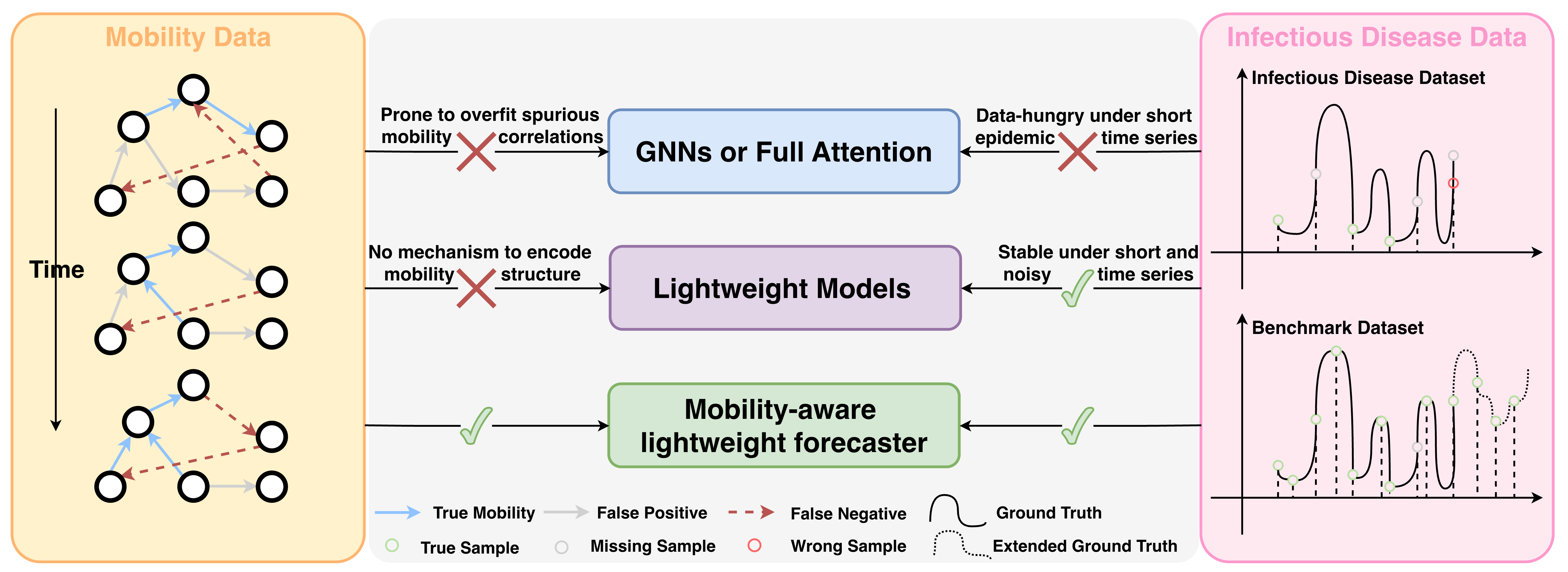}
    \caption{Modeling challenges under real-world epidemic data.}
    \label{fig:current_challenge}
\end{figure}

These data characteristics pose a challenge for modern multivariate time-series forecasters.  Mobility datasets provide only indirect~\cite{gataloAssociationsPhoneMobility2021} and noisy~\cite{sillsAggregatedMobilityData2020,kraemerEffectHumanMobility2020} proxies for transmission, while regional case series are typically short and reported at coarse temporal resolution. Together, these conditions limit the reliability of purely data-driven dependency learning from both mobility and case data. 
As a result, models that perform well on benchmark datasets, where long and clean sequences allow inter-series relationships to be learned directly, often struggle in epidemic forecasting settings~\cite{yuSpatiotemporalGraphConvolutional2018,liDiffusionConvolutionalRecurrent2018,liuSTAEformerSpatiotemporalAdaptive2023}. Forecasters with heavy relational components, such as graph neural networks (GNNs) or full attention mechanisms, are prone to overfitting spurious spatial or temporal patterns. In contrast, lightweight forecasters are typically more robust but lack sufficient capacity to recover meaningful cross-regional dependencies when these dependencies originate from imperfect external drivers such as mobility. 
As shown in Figure \ref{fig:current_challenge}, this tension raises a core question: \textbf{How can lightweight forecasters leverage external relational structure without heavy architectural modifications?}

These challenges have motivated a wide range of epidemic forecasting approaches, spanning mechanistic, statistical, and deep learning models. 
Mechanistic models, such as SIR and SEIR, explicitly encode disease dynamics through compartmental equations and can incorporate inter-regional coupling, but rely on strong assumptions about homogeneous mixing and contact rates and scale poorly with increasing spatial granularity~\cite{hoertelStochasticAgentbasedModel2020,giordanoModellingCOVID19Epidemic2020}. Statistical models, including ARIMA and generalized linear models, offer simplicity and interpretability but have limited capacity to represent nonlinear dynamics or structured spatial dependencies, even when mobility indicators are included as exogenous regressors~\cite{RIBEIRO2020109853}. Deep learning approaches, such as RNNs, LSTMs, and Transformer-based forecasters, can model complex temporal patterns and integrate auxiliary signals like mobility, demographics, or weather, but typically treat mobility as an unstructured correlational input rather than a directional driver of disease spread~\cite{CHIMMULA2020109864}. Spatio-temporal deep learning models extend these ideas by explicitly modeling regional interactions, often using mobility-derived graphs, and have demonstrated strong empirical performance in epidemic forecasting~\cite{panagopoulosTransferGraphNeural2021,fritzCombiningGraphNeural2022}. 
However, their reliance on expressive relational components makes them data-hungry and sensitive to noise in public health surveillance and mobility data.

Among existing approaches, deep learning-based forecasters offer the most flexible and scalable framework for modeling high-dimensional epidemic data, but their effectiveness critically depends on how external structure such as mobility is incorporated. A key limitation of existing approaches is that mobility information is typically incorporated as a correlational signal rather than a structured causal guide, while the low-data and noisy nature of epidemic time series is often overlooked. Our design intuition is that mobility information should act as a soft guide rather than a hard constraint. Learning spatial dependencies entirely from data is akin to searching for structure without a map, while enforcing mobility patterns rigidly risks following an inaccurate one. Instead, a forecasting model should consult mobility-derived structure when it is informative and rely on data-driven temporal patterns when they are reliable. This motivates our approach: we integrate a causal mobility prior through a learnable gating mechanism that adaptively controls its influence, enabling lightweight forecasters to exploit external relational structure without sacrificing robustness in noisy and data-limited settings.

To address these challenges, we propose the Mobility-Informed Causal Adapter (MiCA), a lightweight module that integrates mobility-derived causal structure into temporal forecasting models. MiCA infers directed mobility relations through causal discovery.
It incorporates these relations via gated residual mixing, allowing models to selectively exploit meaningful spatial dependencies while remaining robust to noise and limited data. Our key contributions are as follows:

\begin{itemize}
\item \textbf{Mobility-Informed Causal Adapter (MiCA):}
We introduce MiCA that injects causally inferred mobility structure into time series forecasters without relying on attention or GNN layers. MiCA transforms a causally-derived mobility map into a directional, magnitude-aware prior and integrates it through lightweight linear mixing, enabling effective use of external relational structure in low-data and noisy epidemic settings.

\item \textbf{Adaptive Gating and Leakage-Free Causal Integration:}  
MiCA employs a two-level gating mechanism to control the influence of the causal mobility prior. A global gate regulates overall reliance on mobility information, while an edge-wise gate suppresses spurious links, preventing overfitting to noisy mobility artifacts. The causal prior is constructed exclusively from historical mobility data within the training window, ensuring leakage-free integration and reliable inference at test time.

\item \textbf{Plug-and-Play Design for Lightweight Forecasters:}
MiCA is architecture-agnostic and can be attached to a wide range of temporal backbones without modifying their internal design. We demonstrate its effectiveness on two representative lightweight forecasters, DLinear and a RAM-pruned PatchTST, showing that MiCA enables mobility-aware spatial reasoning with modest computational overhead. This plug-and-play property broadens the applicability of modern deep forecasters to epidemiological settings.

\item \textbf{Comprehensive Empirical Validation:}
We evaluate MiCA on four real-world infectious disease datasets, COVID-19 Incidence, COVID-19 Mortality, Influenza, and Dengue Fever. Across datasets and forecasting horizons, MiCA consistently reduce the average relative error of lightweight backbones by 7.5\%.
\end{itemize}

\section{Related Work}

\subsection{Spatio-Temporal Forecasting}
Spatio-temporal deep learning models were initially developed for traffic forecasting, where graph neural networks were used to capture spatial dependencies among sensors and recurrent or convolutional models were used to encode temporal dynamics~\cite{liDiffusionConvolutionalRecurrent2018,yuSpatiotemporalGraphConvolutional2018}. 
Subsequent work showed that spatial relationships are not purely distance-based, leading to data-driven graph learning approaches that employ learnable node embeddings or attention mechanisms to model adaptive spatial structure~\cite{wuGraphWaveNetDeep2019,guoLearningDynamicsHeterogeneity2022,liuSTAEformerSpatiotemporalAdaptive2023}.

These models have been adapted to epidemic forecasting, where human mobility is used to define inter-regional interactions. 
Prior work has combined graph neural networks with recurrent or Transformer-based temporal encoders for COVID-19 and influenza forecasting~\cite{fritzCombiningGraphNeural2022,panagopoulosTransferGraphNeural2021,gaoSTANSpatiotemporalAttention2021,dengColaGNNCrosslocationAttention2020}. 
While effective in capturing spatial correlations, most existing approaches rely on heuristic or correlation-based mobility representations and require expressive relational modules that are sensitive to noise and data scarcity, limiting robustness in public health surveillance settings.

\subsection{Causality-Aware Time-Series Modeling}
Causal discovery methods aim to recover directional relationships among time-dependent variables from observational data, including constraint-based approaches such as PC, FCI, and PCMCI~\cite{spirtesAlgorithmFastRecovery1991,spirtesCausalInferencePresence1995,rungeDetectingQuantifyingCausal2019}, asymmetry-based methods such as Granger causality and LiNGAM~\cite{grangerInvestigatingCausalRelations1969,hyvarinenEstimationStructuralVector2010}, and score-based methods including NOTEARS and its time-series extensions~\cite{pamfilDYNOTEARSStructureLearning2020}. 
These methods provide principled tools for inferring causal structure but are typically studied independently of downstream forecasting objectives~\cite{rungeCausalInferenceTime2023,rungeInferringCausationTime2019}.

More recent work has explored integrating causal reasoning into forecasting architectures. TCDF, Neural Granger Causality, and CR-VAE encode causal sparsity or directionality within neural models~\cite{nautaCausalDiscoveryAttentionbased2019,tankNeuralGrangerCausality2022,liCausalRecurrentVariational2023}, while CausalFormer and CausalGNN incorporate causal structure into Transformer- and GNN-based forecasting frameworks~\cite{zhangCausalformerCausalDiscoverybased2023,wangCausalGNNCausalbasedGraph2022}. 
Related ideas have also been explored in attention regularization through causal intervention~\cite{yangCausalAttentionVisionlanguage2021}. 
Despite these advances, existing approaches typically rely on static or implicitly learned causal structure and focus on parameter-heavy architectures, leaving open how to integrate empirically inferred, mobility-driven causal relations into lightweight forecasters in a data-efficient manner.

\section{Methods}

\begin{figure*}[t]
    \centering
    \includegraphics[width=0.9\textwidth]{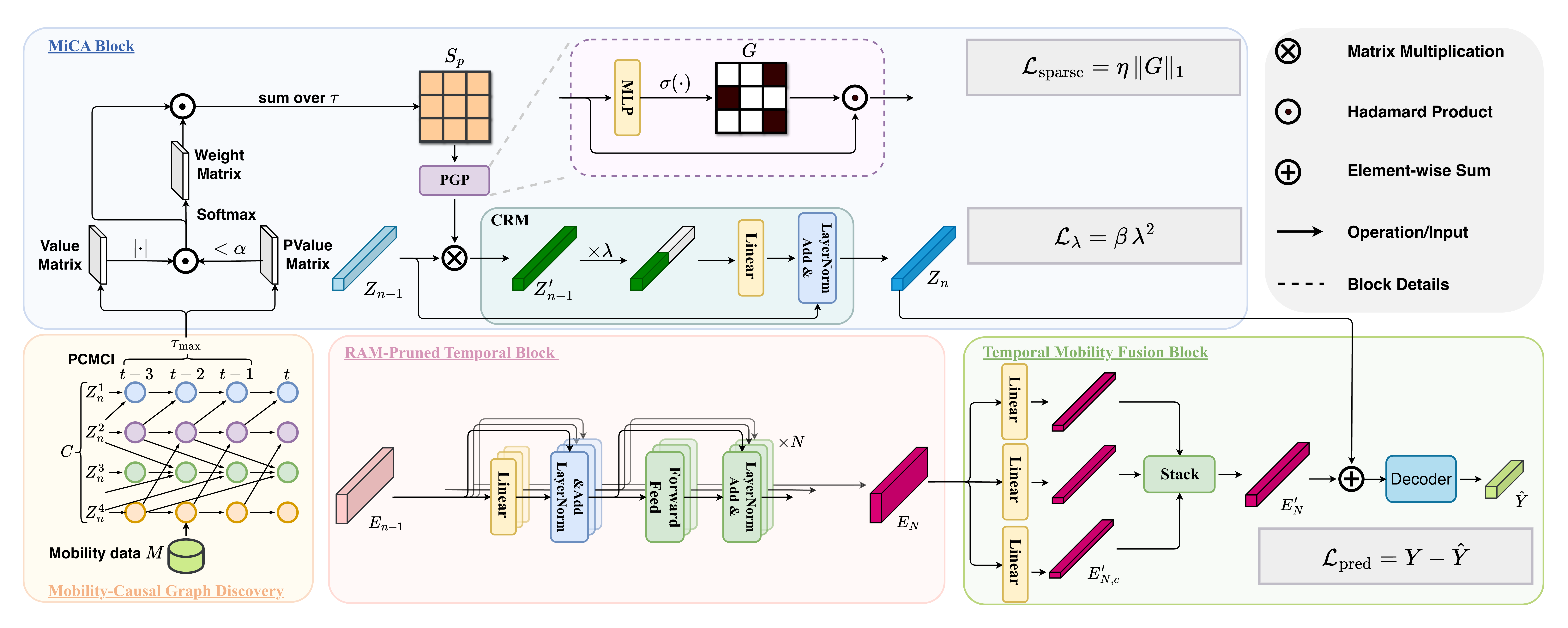}
    \caption{End-to-end MiCA pipeline. PCMCI derives causal mobility relations, which MiCA incorporates using link-level gating (PGP) and gated residual mixing (CRM). The model fuses mobility-aware spatial features with a RAM-pruned temporal encoder, trained under prediction and prior-regularization losses.}
    \label{fig:main_fig}
\end{figure*}

\subsection{Preliminary}
In multivariate time series forecasting, the goal is to predict future values $\mathrm{\hat{Y}} = \{\mathrm{x}_{L+1}, \dots, \mathrm{x}_{L+T}\} \in \mathbb{R}^{T \times C}$ given past observations $\mathrm{X} = \{\mathrm{x}_1, \dots, \mathrm{x_L}\} \in \mathbb{R}^{L \times C}$, where $L$ is the lookback window and $C$ is the number of channels. Let $\mathcal{F}(\cdot;\theta)$ denote a forecasting model parameterized by $\theta$. The prediction task is formulated as:
\begin{align}
    \mathrm{\hat{Y}} = \mathcal{F}(X; \theta).
\end{align}

\subsection{Transformer-based Temporal Framework}
Transformer-based forecasters employ multi-head attention (MHA) to capture dependencies across time steps and channels. For models such as PatchTST~\cite{nieTimeSeriesWorth2023}, the input sequence $\mathbf{X}$ is divided into $N$ patches of length $P$ through:
\begin{equation}
    X_p = \text{Patching}(X),
\end{equation}
where $X_p \in \mathbb{R}^{N \times (P \times C)}$ represents the patches. Each patch is embedded into a latent space with positional encoding:
\begin{equation}
    E_0 = \text{Embedding}(X_p),
\end{equation}
with $E_0 \in \mathbb{R}^{N \times d_{\text{head}}}$. For clarity, we omit the explicit decomposition of embeddings into multiple attention heads. 

\subsubsection{Multi-Head Attention (MHA)}

Let $E_{n-1}^{(h)} \in \mathbb{R}^{{N} \times d_{\text{head}}}$ denote the input to attention head $h \in \{1, \dots, H\}$. Each head projects its input into query, key, and value representations:
\begin{equation}
    \resizebox{.89\linewidth}{!}{$
        \displaystyle
        Q^{(h)} = E_{n-1}^{(h)} W_q^\top, K^{(h)} = E_{n-1}^{(h)} W_k^\top, V^{(h)} = E_{n-1}^{(h)} W_v^\top
    $},
\end{equation}
where $W_{ \{ q, k, v \} } \in \mathbb{R}^{d_{\text{head}} \times d_{\text{head}}}$ are learnanble parameters. The raw attention scores $A \in \mathbb{R}^{N \times N}$ are computed using scaled dot-product attention:
\begin{equation}
    A^{(h)} = \frac{Q^{(h)}(K^{(h)})^T}{\sqrt{d_{\text{head}}}}.
\end{equation}
followed by a softmax to get attention score matrix:
\begin{equation}
    S^{(h)} = \text{Softmax}(A^{(h)}).
\end{equation}
The final output $O^{(h)} \in \mathbb{R}^{N \times d_{\text{head}}} $ is given by:
\begin{align}
    O^{(h)} &= S^{(h)}V^{(h)}.
\end{align}
Outputs from all heads are concatenated and linearly projected:
\begin{align}
    E_{n}^{\prime} = \left[ O^{(1)}, \dots, O^{(H)} \right] W_E. \label{eq:attention_output}
\end{align}
where $W_E \in \mathbb{R}^{d_{\text{model}} \times d_{\text{model}}}$. We summarize the multi-head attention operation as
\begin{align}
    E_{n}^{\prime} = \text{MHA}(E_{n-1}). \label{eq:multihead}
\end{align}

\subsubsection{Full Attention Block}

A standard Transformer layer augments the attention mechanism with residual connections, layer normalization, dropout, and a position-wise feedforward network. Formally, the block is defined as:
\begin{align} 
    R_n &= \text{LayerNorm}(\text{Dropout}(\text{MHA}(E_{n-1})) + E_{n-1}), \label{eq:ln1} \\
    F_n &= \sigma(R_nW_F)W_B, \label{eq:ffn} \\ 
    E_n &= \text{LayerNorm}(\text{Dropout}(F_n) + R_n).\label{eq:ln2} 
\end{align} 
$W_F \in \mathbb{R}^{d_{\text{model}} \times d_{\text{ff}}}$ and $W_B \in \mathbb{R}^{d_{\text{ff}} \times d_{\text{model}}}$ are feedforward projection matrices, and $\sigma(\cdot)$ denotes the activation funciton. For simplicity, we summarize the Eq. \eqref{eq:ln1} - \eqref{eq:ln2} as a single attention block:
\begin{align}
    E_n = \text{AttentionBlock}(E_{n-1}).
\end{align}
Stacking $B$ such blocks yields the full temporal encoder:
\begin{equation} \label{eq:attentionblock}
    E_N = \mathcal{O}_{n=1}^{B} \oplus(\text{AttentionBlock}_n(E_{n-1})),
\end{equation}
where the $\mathcal{O}_{n=1}^{B}$ represents a composition of functions over $B$ layers.

\subsubsection{RAM-Pruned Temporal Backbone}

Following the Replace Attention with MLP (RAM) rationale~\cite{guo2025ramreplaceattentionmlp}, we remove MHA from each block to avoid overfitting on short, noisy epidemic time series.  
Eliminating $\text{MHA}(\cdot)$ from the $\text{AttentionBlock}$ in Eq. \eqref{eq:ln1} - \eqref{eq:ln2} yields a RNF block:
\begin{equation}
    E_n = \text{RNF}(E_{n-1}),
\end{equation}
which retains residual connections, layer normalization, and the feedforward sublayer, while omitting attention entirely. Stacking $B$ RNF blocks forms the RAM-pruned temporal encoder:
\begin{equation}
    E_N = \mathcal{O}_{n=1}^{B} \oplus\big(\text{RNF}_n(E_{n-1})\big).
\end{equation}
This lightweight backbone preserves effective temporal modeling capacity while mitigating overfitting in data-limited epidemic forecasting settings. Its simplicity and efficiency also make it a natural foundation for integrating the proposed MiCA module.

\subsection{Mobility-Causal Graph Discovery}
To capture inter-regional dependencies induced by human movement, we construct a causal mobility graph from historical mobility data. Let
\begin{equation}
    M = \{m_{u,c}\}_{u=1,c=1}^{U,C} \in \mathbb{R}^{U \times C},
\end{equation}
denote the mobility time series, where $C$ is the number of regions and $U$ is the number of time steps used for causal discovery. To prevent information leakage, mobility data are restricted to observations prior to the end of the training period. The resulting causal prior remains fixed during evaluation.

Causal discovery requires weak stationarity to avoid spurious dependencies. We apply first-order differencing followed by z-score normalization:
\begin{equation}
    \tilde{m}_{u,c} = \frac{(m_{u,c} - m_{u-1,c}) - \mu_c}{\sigma_c},
\end{equation}
where $\mu_c$ and $\sigma_c$ are computed from the differenced series of region $c$. This preprocessing removes global trends and aligns scales across regions.

We apply the PCMCI (Peter–Clark Momentary Conditional Independence) algorithm~\cite{rungeDetectingQuantifyingCausal2019}, which is designed for high-dimensional and autocorrelated time series. PCMCI combines PC-based skeleton discovery with a Momentary Conditional Independence (MCI) test to identify directed lagged dependencies. In our implementation, conditional independence is assessed using partial correlation (ParCorr). PCMCI proceeds in two stages:
\begin{enumerate}
    \item \textbf{PC step}: iteratively removes non-significant lagged edges using conditional independence tests;
    \item \textbf{MCI step}: re-evaluates retained edges while conditioning on discovered parents to compute final test statistics.
\end{enumerate}
Algorithmic details are provided in supplementary 

Given the preprocessed mobility data $\tilde{M}$, the PCMCI returns:
\begin{equation}
    [\text{Val}, \text{Pval}] = \text{PCMCI}_{\text{ParCorr}}(\tilde{M}; \tau_{\max}-1),
\end{equation}
where $\text{Val} \in \mathbb{R}^{C \times C \times \tau_{\max}}$ encodes signed causal strengths and $\text{Pval} \in \mathbb{R}^{C \times C \times \tau_{\max}}$ stores corresponding significance values. Each slice $\text{Val}_{:,:, \tau}$ represents a directed mobility graph at lag $\tau$.

\subsection{Mobility-Informed Causal Adapter (MiCA)}
Using the causal discovery outputs $\text{Val}$ and $\text{Pval}$, we construct a mobility-informed causal prior matrix $S_p \in \mathbb{R}^{C \times C}$ that summarizes directed inter-regional dependencies. The aggregated causal strength from region $j$ to region $i$ is defined as:
\begin{equation}
S_{p,ij} = \sum_{\tau = 1}^{\tau_{\max}} w_{i, j,\tau} \mathbb{I}\left[ \text{Pval}_{i,j,\tau} < \alpha \right] \left| \text{Val}_{i,j,\tau} \right|,
\label{eq:Sp_sum}
\end{equation}
where $\mathbb{I}[\cdot]$ denotes the indicator function, $\alpha$ is the significance threshold (set to be $0.05$), and $w_{i,j,\tau}$ weights the contribution of different time lags.

To emphasize stronger causal effects, lag weights are normalized using an exponential decay kernel:
\begin{equation}
w_{i, j,\tau} =
\frac{\exp(-\text{Val}_{i,j,\tau} / \kappa)}
{\sum_{\nu = 1}^{\tau_{\max}} \exp(-\text{Val}_{i,j,\nu} / \kappa)},
\quad
\text{s.t.} \quad
\sum_{\tau = 1}^{\tau_{\max}} \text{Val}_{i,j,\tau} = 1,
\label{eq:lag_weight}
\end{equation}
where $\kappa>0$ controls the sharpness of the weighting distribution and is set to $1$ in our experiments. This design amplifies dominant causal links while suppressing weak or noisy associations.

The resulting matrix $S_p$ encodes a directed, magnitude-aware mobility prior that reflects both statistical significance and causal strength. This prior is subsequently integrated into the forecasting model to guide spatial information flow and improve predictive performance. Let $Z_{n-1}\in\mathbb{R}^{C\times d_{\text{model}}}$ denote the spatial feature input at layer $n$.

\subsubsection{Causal Residual Mixer (CRM)}

The Causal Residual Mixer (CRM) injects mobility-induced causal information through linear propagation:
\begin{equation}
    Z_{n-1}^{\prime} \;=\; S_p\, Z_{n-1}.
\end{equation}
The propagated causal message is integrated using a gated residual connection:
\begin{equation}
    Z_n \;=\; \mathrm{LayerNorm}\!\big(Z_{n-1} \;+\; W_o\,(\lambda Z_{n-1}^{\prime})\big),
\end{equation}
where $W_o\!\in\!\mathbb{R}^{d_{\text{model}}\times d_{\text{model}}}$ is a learnable projection matrix and $\lambda\ge 0$ controls the strength of causal mixing.
To ensure stable optimization, $\lambda$ is parameterized using a softplus function:
\begin{equation}
\lambda = \text{softplus}(\theta),\quad 
\text{softplus}(\theta)= \ln(1+e^{\theta}),
\end{equation}
where $\theta$ is unconstrained. Initializing $\theta < 0$ ensures the model begins with minimal causal influence and increases reliance on $S_p$ only when beneficial.

\subsubsection{Prior-Gated Projection (PGP)}
Complementing the global mixing gate in CRM, the Prior-Gated Projection (PGP) introduces edge-wise gating to suppress unreliable causal links. Specifically, an adaptive gate matrix is computed as
\begin{equation}
    G = \sigma(\text{MLP}(S_p)).
\end{equation}
where $G\in\mathbb{R}^{C\times C}$ assigns an adaptive confidence to each entry of $S_p$. The gated causal propagation is:
\begin{equation}
    Z_{n-1}^{\prime} \;=\; (G \odot S_p)\, Z_{n-1}, 
\end{equation}
followed by the same residual fusion as in CRM. PGP therefore refines the causal graph by suppressing noisy edges while preserving informative links.

\subsubsection{Adpater Loss Function}
In addition to the prediction loss, MiCA introduces two lightweight regularizers:
\begin{equation}
    \mathcal{L}=\mathcal{L}_{\text{pred}}+\mathcal{L}_\lambda + \mathcal{L}_{\text{sparse}}.
\end{equation}
To prevent excessive reliance on the prior, we penalize large global mixing weights:
\begin{equation}
    \mathcal{L}_\lambda=\beta\,\lambda^2,
\end{equation}
where $\beta$ controls the strength of regularization. To encourage sparsity in edge-wise gating, we apply an $\ell_1$ regularizer:
\begin{equation}
    \mathcal{L}_{\text{sparse}}=\eta\,\|G\|_1,
\end{equation}
where $\eta$ controls sparsity strength. Given the limited data available in epidemic forecasting, we use a single spatial adapter layer ($D = 1$) to reduce overfitting:
\begin{equation}
    Z_N = \mathcal{O}_{n=1}^{D} \oplus(\text{MiCA}_n(Z_{n-1})).
\end{equation}

The gated residual design also provides a simple stability guarantee: MiCA induces a bounded, controllable perturbation to the backbone representation.

\begin{theorem}[Bounded Influence of MiCA]
Let $\mathcal{F}_\theta$ denote the forecasting network after the MiCA injection point, viewed as a mapping from the fused representation to the output. Assume $\mathcal{F}_\theta$ is $L$-Lipschitz under the Frobenius norm. 
For a MiCA update of the form $\tilde{Z} = Z + W_o(\lambda \, \bar{S} Z)$, where $\lambda \ge 0$, $W_o$ is a learnable projection, and $\bar{S}$ is the (possibly gated) mobility prior (i.e., $\bar{S}=S_p$ for CRM and $\bar{S}=G\odot S_p$ for PGP), the output perturbation satisfies
\begin{equation}
\|\mathcal{F}_\theta(\tilde{Z}) - \mathcal{F}_\theta(Z)\|_F 
\;\le\;
L\,\lambda\,\|W_o\|_2\,\|\bar{S}\|_2\,\|Z\|_F.
\label{eq:end2end_bound}
\end{equation}
In particular, the influence of MiCA is explicitly controlled by the global mixing weight $\lambda$ and the spectral norm of the mobility prior.
\end{theorem}
\noindent\textit{Proof sketch.} By Lipschitz continuity of $\mathcal{F}_\theta$ and submultiplicativity of induced norms; detailed bounds for CRM/PGP are provided in Supplementary 

\subsection{Temporal-Mobility Fusion}
Let $E_N$ and $Z_N$ denote the final outputs of the temporal backbone and the MiCA spatial adapter, respectively. We fuse them using a channel-wise linear projection followed by element-wise addition. For each region $c \in {1,\dots,C}$, we compute
\begin{equation}
    E_{N,c}^{\prime}=W_cE_N, 
\end{equation}
where $E_{N,c}^{\prime}\in\mathbb{R}^{d_{\text{model}}}$. The projected features are stacked across regions:
\begin{equation}
    E_N^{\prime}=\text{Stack}\big(\{E_{N,c}^{\prime}\}_{c=1}^C\big),
\end{equation}
yielding $E_N^{\prime}\in\mathbb{R}^{C\times d_{\text{model}}}$.
The integrated representation is then:
\begin{equation}
    O_N = E_N^{\prime} + Z_N.
\end{equation}
Finally, the forecasting output is obtained via a decoder:
\begin{equation}
    \hat{Y} = \text{Decoder}(O_N),
\end{equation}
implemented as a linear projection in practice. This design preserves a one-to-one alignment between temporal and spatial features at the channel level while enabling additive interaction between the two branches.

\section{Experiments}
Experiments are conducted on a machine equipped with NVIDIA GeForce RTX 2080 GPU, each with 16GB memory. The operating system is Ubuntu 18.04, and we use Pytorch version 2.6.0, and Python version 3.10.4 for all experiments. 

We evaluate MiCA on four real-world epidemic datasets covering daily and weekly surveillance with diverse spatial structures: COVID-19 incidence and mortality (daily, 49 U.S. states), Influenza (weekly, 49 U.S. states), and Dengue Fever (weekly, 27 Brazilian municipalities). Across all datasets, we adopt a sliding-window multivariate forecasting setup with a lookback window of $7$ time steps and forecasting horizons of $7, 14, 21, 28$ steps ahead. Each dataset is split into training, validation, and test sets with a ratio of $0.6:0.2:0.2$.

We evaluate MiCA by integrating it into two lightweight temporal backbones, including a RAM-pruned PatchTST (abbreviated as RAM in tables for brevity) and DLinear. We compare them against (i) classical statistical models (AR, ARMA~\cite{contrerasARIMAModelsPredict2003}), (ii) recurrent neural models (vanilla RNNs~\cite{werbosBackpropagationTimeWhat1990}, GRUs~\cite{choLearningPhraseRepresentations2014}, LSTMs~\cite{hochreiterLongShorttermMemory1997}), and (iii) spatio-temporal models that explicitly model spatial interactions (DCRNN~\cite{liDiffusionConvolutionalRecurrent2018}, STGCN~\cite{yuSpatiotemporalGraphConvolutional2018}, ColaGNN~\cite{dengColaGNNCrosslocationAttention2020}).

\subsection{Results}

\begin{table*}[t]
\centering
\resizebox{\textwidth}{!}{\begin{tabular}{{@{}lrrrrrrrr@{}}}
\toprule
 & \multicolumn{4}{c}{RMSE} & \multicolumn{4}{c}{MAE} \\ 
\cmidrule(lr){2-5} \cmidrule(lr){6-9}  
Dataset & COVID-Incidence & COVID-Mortality & Influenza & Dengue & COVID-Incidence & COVID-Mortality & Influenza & Dengue \\
\midrule
AR & $1.069\pm0.055$ & $1.978\pm0.139$ & $9.734\pm0.148$ & $95.785\pm2.858$ & $0.755\pm0.042$ & $1.389\pm0.132$ & $7.023\pm0.189$ & $34.259\pm1.731$ \\
ARMA & $1.145\pm0.189$ & $3.667\pm1.836$ & $13.476\pm6.616$ & $99.385\pm22.941$ & $0.847\pm0.153$ & $3.009\pm1.459$ & $9.992\pm5.025$ & $39.438\pm17.534$ \\
RNN & $1.152\pm0.002$ & $1.594\pm0.012$ & \underline {$8.431\pm0.135$} & $94.816\pm0.594$ & $0.846\pm0.003$ & $1.001\pm0.020$ & $5.658\pm0.153$ & $34.892\pm0.841$ \\
GRU & $1.119\pm0.010$ & $1.576\pm0.024$ & $8.713\pm0.214$ & $94.488\pm0.465$ & $0.812\pm0.005$ & $0.983\pm0.026$ & $5.865\pm0.124$ & $34.268\pm1.237$ \\
LSTM & $1.217\pm0.048$ & $1.640\pm0.045$ & $9.216\pm0.096$ & $94.731\pm0.447$ & $0.908\pm0.047$ & $1.087\pm0.052$ & $6.398\pm0.085$ & $32.542\pm0.325$ \\
DCRNN & $1.226\pm0.026$ & $2.409\pm0.165$ & $9.504\pm0.069$ & $95.253\pm0.460$ & $0.924\pm0.014$ & $2.037\pm0.189$ & $6.884\pm0.188$ & $32.306\pm0.703$ \\
STGCN & $1.177\pm0.042$ & $1.514\pm0.007$ & $8.693\pm0.671$ & $119.201\pm9.780$ & $0.872\pm0.042$ & $0.908\pm0.011$ & $6.015\pm0.548$ & $49.285\pm3.889$ \\
ColaGNN & $1.151\pm0.050$ & $1.654\pm0.048$ & $9.095\pm0.223$ & $97.746\pm5.506$ & $0.856\pm0.051$ & $1.177\pm0.083$ & $6.316\pm0.227$ & $36.605\pm3.909$ \\ \midrule
RAM & \underline{$0.941\pm0.006$} & \underline{$1.305\pm0.004$} & $8.896\pm0.168$ & \underline{$70.530\pm0.013$} & \underline{$0.499\pm0.003$} & \underline{$0.534\pm0.002$} & $5.833\pm0.248$ & \underline{$21.290\pm0.029$} \\
RAM + MiCA & \bm{$0.934\pm0.010$} & \bm{$1.257\pm0.005$} & $8.526\pm0.027$ & \bm{$70.365\pm0.105$} & \bm{$0.498\pm0.003$} & \bm{$0.516\pm0.002$} & \bm{$5.327\pm0.062$} & \bm{$21.213\pm0.194$} \\
DLinear & $1.055\pm0.021$ & $1.877\pm0.043$ & $9.437\pm0.172$ & $92.011\pm1.769$ & $0.783\pm0.057$ & $1.326\pm0.054$ & $6.795\pm0.178$ & $32.423\pm0.644$ \\
DLinear + MiCA & $1.064\pm0.019$ & $1.551\pm0.006$ & \bm{$8.006\pm0.185$} & $87.818\pm0.937$ & $0.761\pm0.017$ & $0.858\pm0.003$ & \underline{$5.582\pm0.154$} & $30.238\pm0.510$ \\ \bottomrule
\end{tabular}}
\caption{Average forecasting performance on four infectious disease datasets under two evaluation metrics (RMSE and MAE). All results are averaged over three independent runs and over four forecasting horizons (7, 14, 21, and 28 steps). \textbf{Bold} values denote the best performance and \underline{underlined} values denote the second-best performance for each dataset. Full results can be found in Supplemntary Table~\ref{s-tab:table_full}.}
\label{tab:main_table}
\end{table*}

Table~\ref{tab:main_table} summarizes the average forecasting performance across four epidemic datasets and multiple forecasting horizons. Overall, integrating MiCA consistently improves lightweight temporal backbones under both RMSE and MAE, confirming the effectiveness of mobility-informed causal integration. The influenza dataset exhibits a slightly different pattern. Due to its weekly reporting granularity, influenza data provide fewer effective observations and are more susceptible to noise, allowing simpler temporal models such as RNNs to perform competitively under RMSE. Nevertheless, RAM+MiCA achieves the best performance under MAE, and DLinear+MiCA ranks second, indicating that mobility-informed causal integration remains effective for stabilizing forecasts in noisy, low-frequency surveillance settings.

For the RAM backbone, MiCA yields uniform performance gains across all datasets, with relative reductions of up to 2.20\% in RMSE and 3.09\% in MAE, and consistent improvements on both daily (COVID-19) and weekly (Influenza, Dengue) surveillance data. These gains allow RAM+MiCA to achieve the best performance on most datasets, outperforming parameter-heavy spatio-temporal models such as DCRNN, STGCN, and ColaGNN. For DLinear, MiCA provides substantial improvements on datasets where spatial dependence is pronounced, including COVID-19 Mortality, Influenza, and Dengue, with relative reductions reaching 12.37\% (RMSE) and 19.96\% (MAE). While RMSE improvements on COVID-19 incidence are marginal, MAE is maintained or improved, indicating no degradation in average forecasting accuracy. Across backbones and datasets, MiCA improves RMSE and MAE by 5.64\% and 9.38\% on average, respectively. These results demonstrate that causal mobility priors can enhance epidemic forecasting accuracy without relying on heavy attention or graph-based spatial modules.


\subsection{Ablation Studies}
We conduct ablation studies to examine the contribution of individual components in MiCA and to validate key design choices. Specifically, we analyze the impact of model structure, forecasting horizon, causal prior construction, and model complexity. 

\begin{figure}[t]
    \centering
    \begin{subfigure}[t]{0.23\textwidth}
        \centering
        \includegraphics[width=\textwidth]{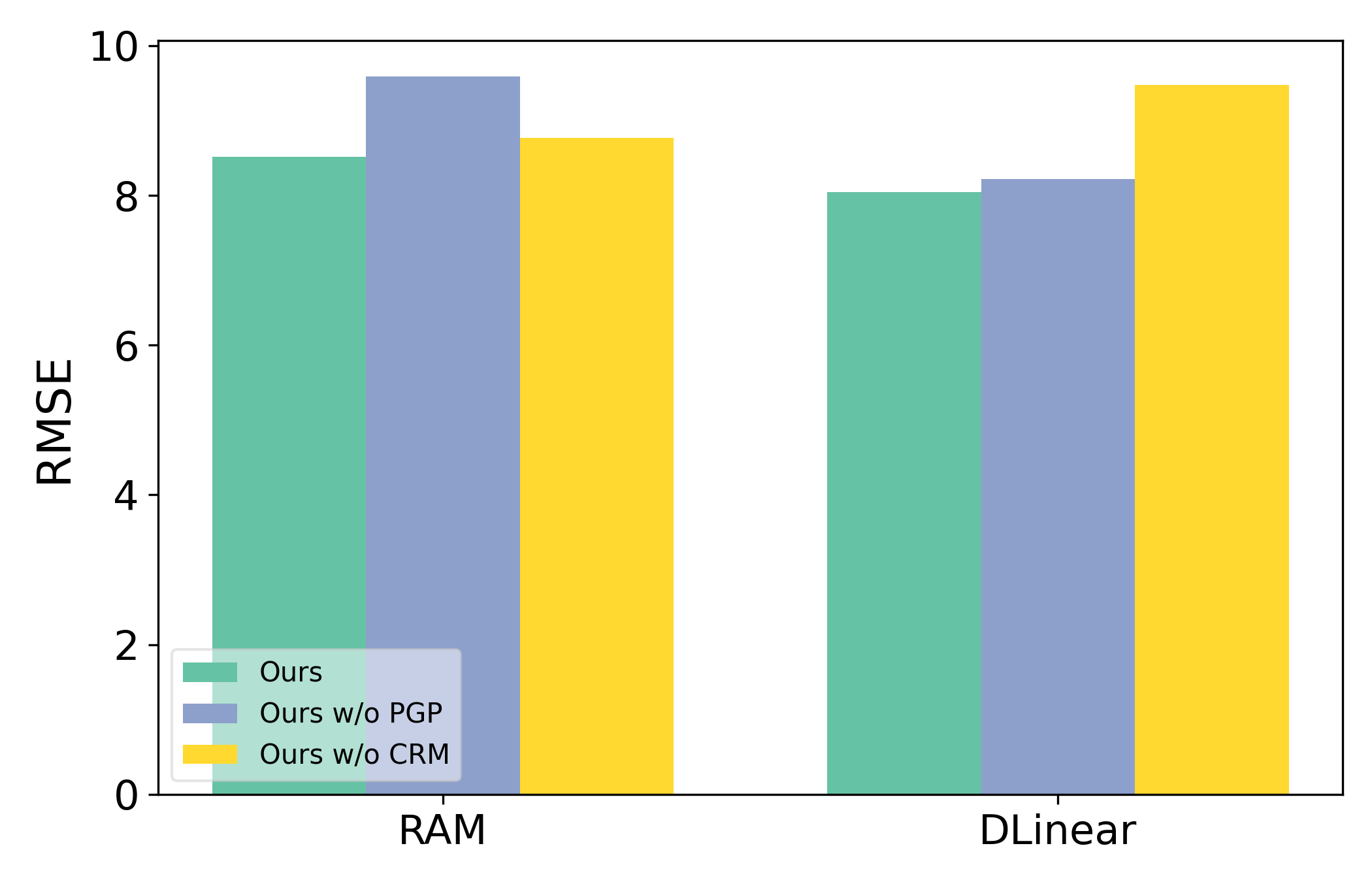}
        \caption{RMSE}
        \label{fig:structure_rmse}
    \end{subfigure}
    \begin{subfigure}[t]{0.23\textwidth}
        \centering
        \includegraphics[width=\textwidth]{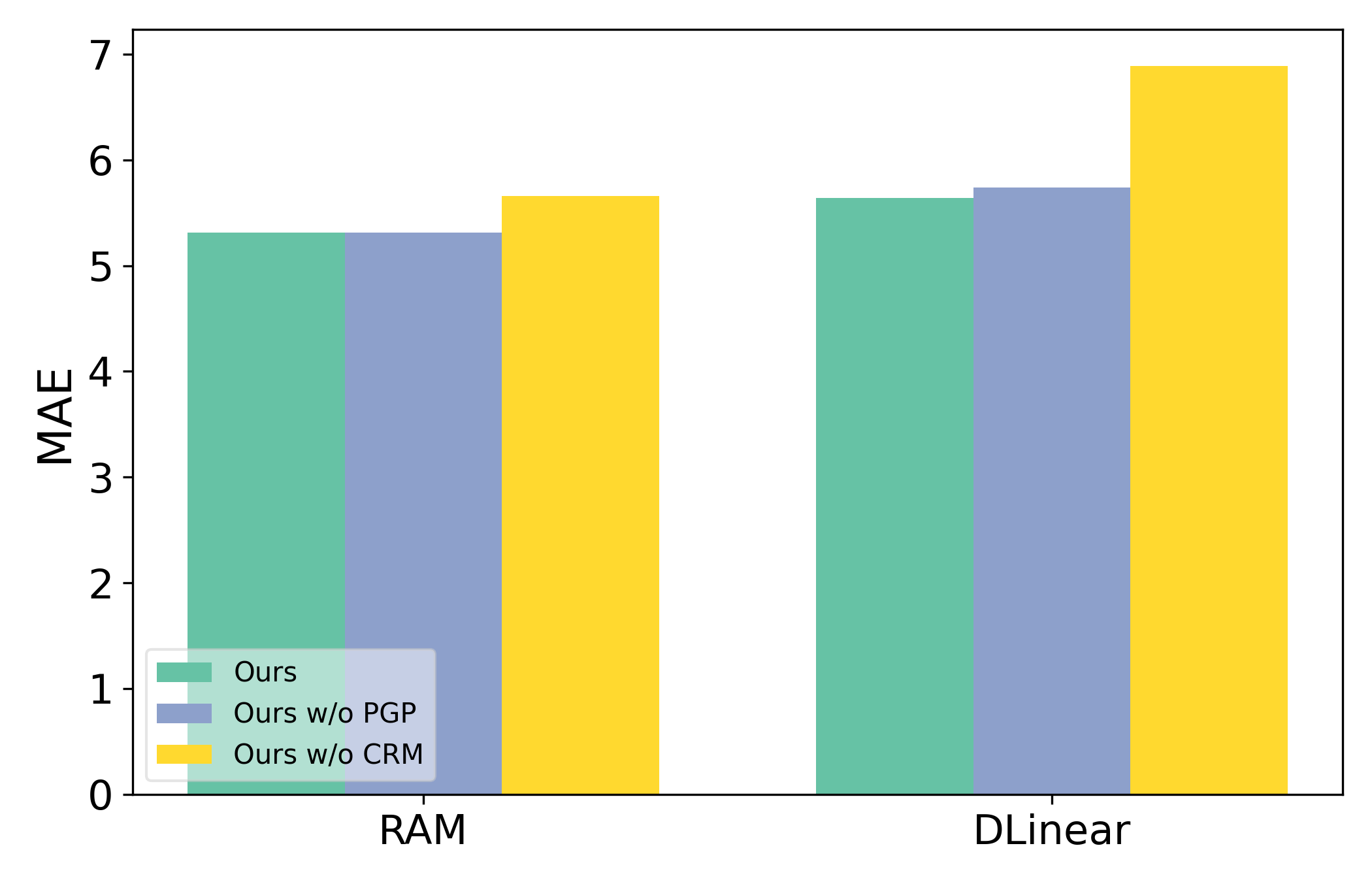}
        \caption{MAE}
        \label{fig:structure_mae}
    \end{subfigure}
  
    \caption{Ablation analysis on major components of the proposed model.}
    \label{fig:abl_structure}
\end{figure}

\subsubsection{Model Structure}
Figure~\ref{fig:abl_structure} examines the contribution of MiCA’s core components by comparing the full model against variants that remove the Prior-Gated Projection (w/o PGP) or the Causal Residual Mixer (w/o CRM). Across both RAM and DLinear backbones, the full MiCA model achieves the lowest RMSE and MAE, indicating that each component contributes to improved forecasting accuracy. Removing PGP consistently degrades performance, suggesting that edge-wise gating is important for filtering noisy or unreliable mobility links. Removing CRM leads to a larger performance drop, particularly for DLinear, highlighting the role of residual causal mixing in injecting mobility-derived structure into lightweight temporal encoders. These results confirm that MiCA’s gains arise from the complementary effects of global residual integration (CRM) and selective edge-level gating (PGP), rather than from either component alone.

\begin{figure}
    \centering
    \includegraphics[width=\linewidth]{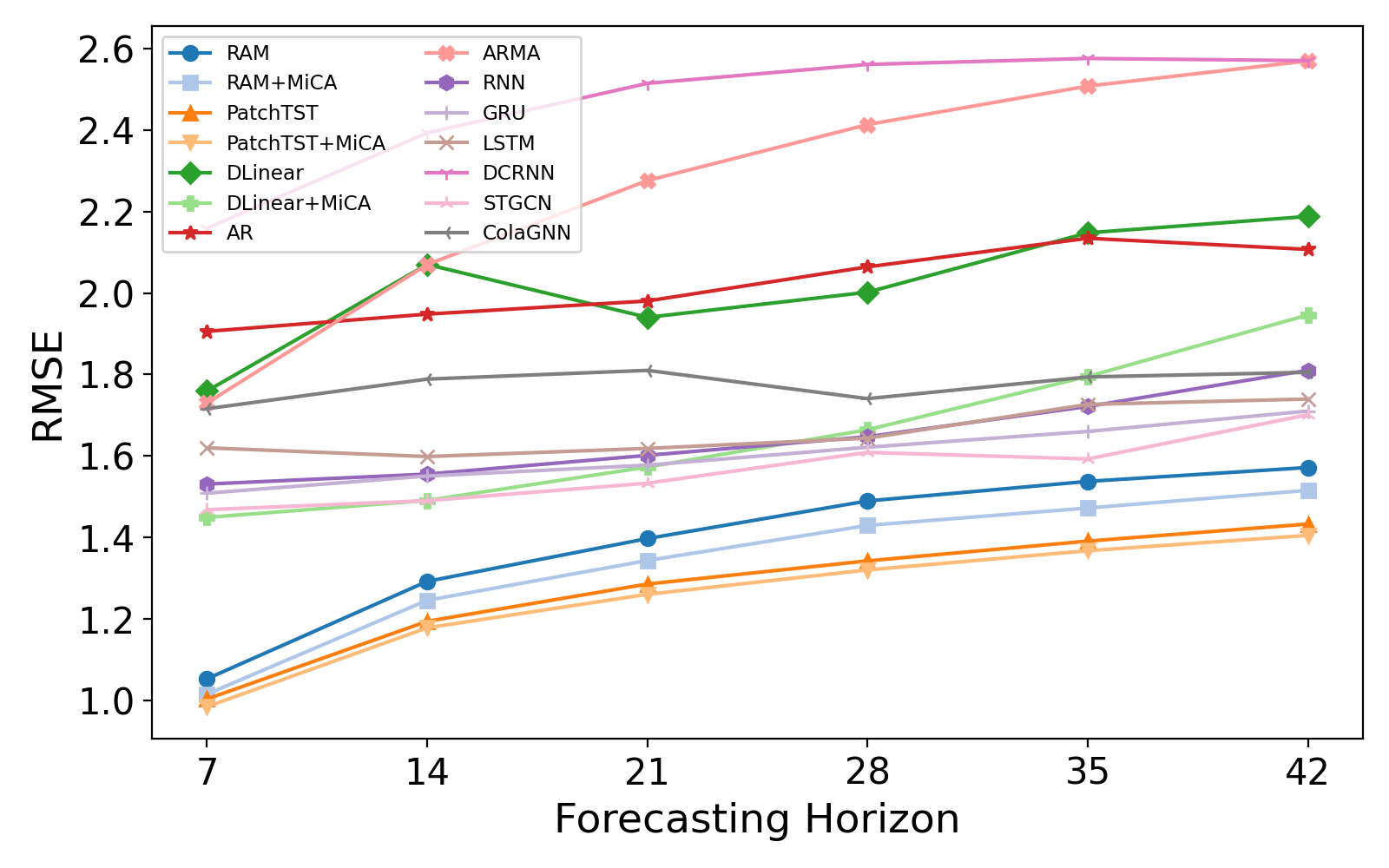}
    \caption{RMSE comparison across forecasting horizons (7–42 steps) on COVID-19 mortality.}
    \label{fig:abl_horizon}
\end{figure}

\subsubsection{Forecasting Horizon}
To assess robustness under longer forecasting horizons, we evaluate models with horizons extended to $\{7,14,21,28,35,42\}$ steps as shown in Figure~\ref{fig:abl_horizon}. As expected, prediction error increases with horizon for all methods, reflecting the inherent difficulty of long-range epidemic forecasting. MiCA consistently improves lightweight backbones, including RAM and DLinear, across all horizons, with gains persisting and in some cases becoming more pronounced at longer prediction ranges. This indicates that mobility-informed causal guidance contributes to stable long-horizon forecasting rather than only short-term accuracy. We also observe consistent gains when applying MiCA to PatchTST (orange lines), although the relative improvement is smaller than for lightweight backbones such as RAM and DLinear. A plausible reason is that PatchTST already provides a strong temporal encoder for each region under the channel-independent setting, so MiCA mainly contributes additional spatial guidance rather than fixing weak temporal modeling.

\begin{figure}[t]
    \centering
    \begin{subfigure}[t]{0.23\textwidth}
        \centering
        \includegraphics[width=\textwidth]{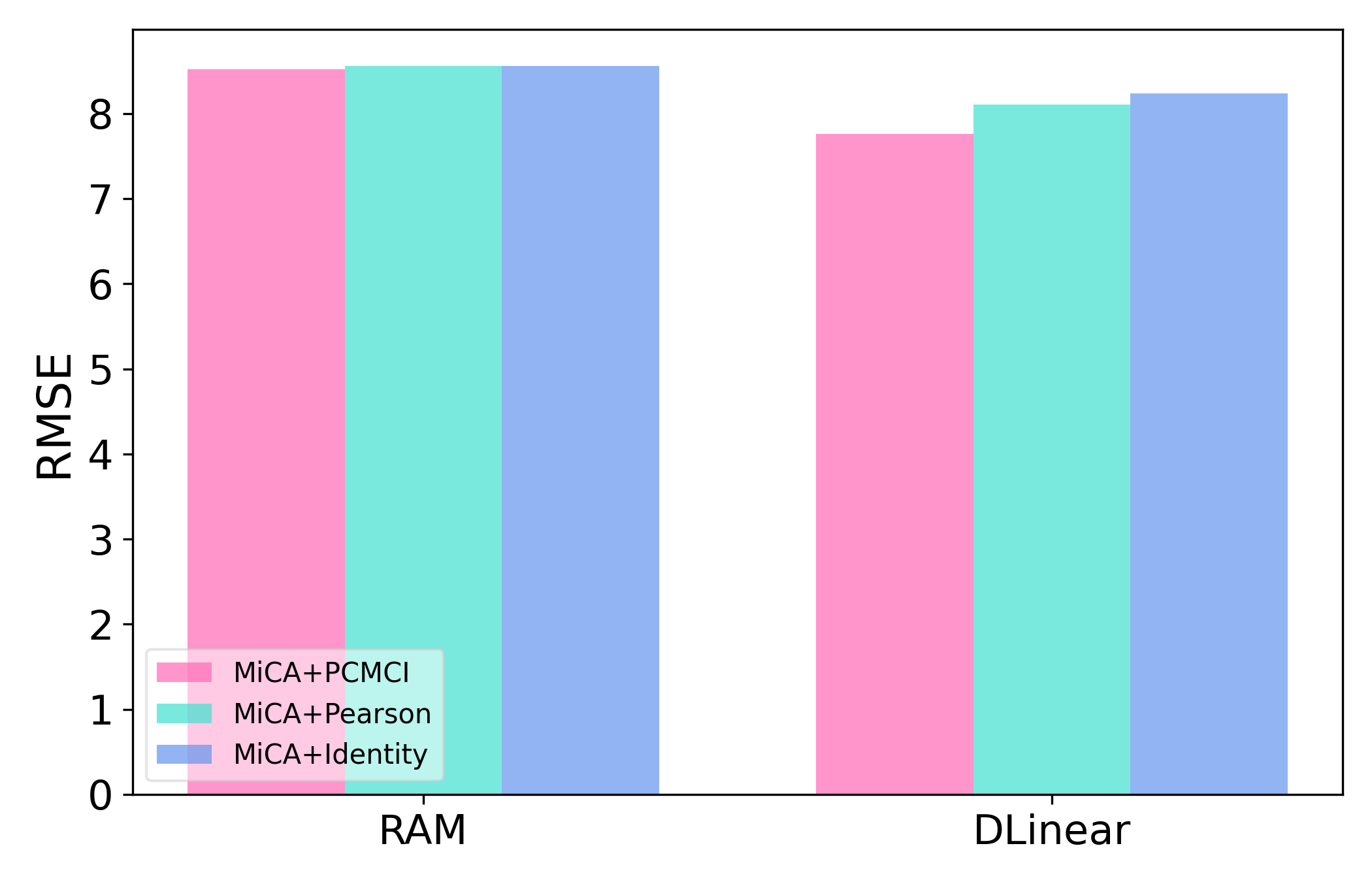}
        \caption{RMSE}
        \label{fig:PCMCI_rmse}
    \end{subfigure}
    \begin{subfigure}[t]{0.23\textwidth}
        \centering
        \includegraphics[width=\textwidth]{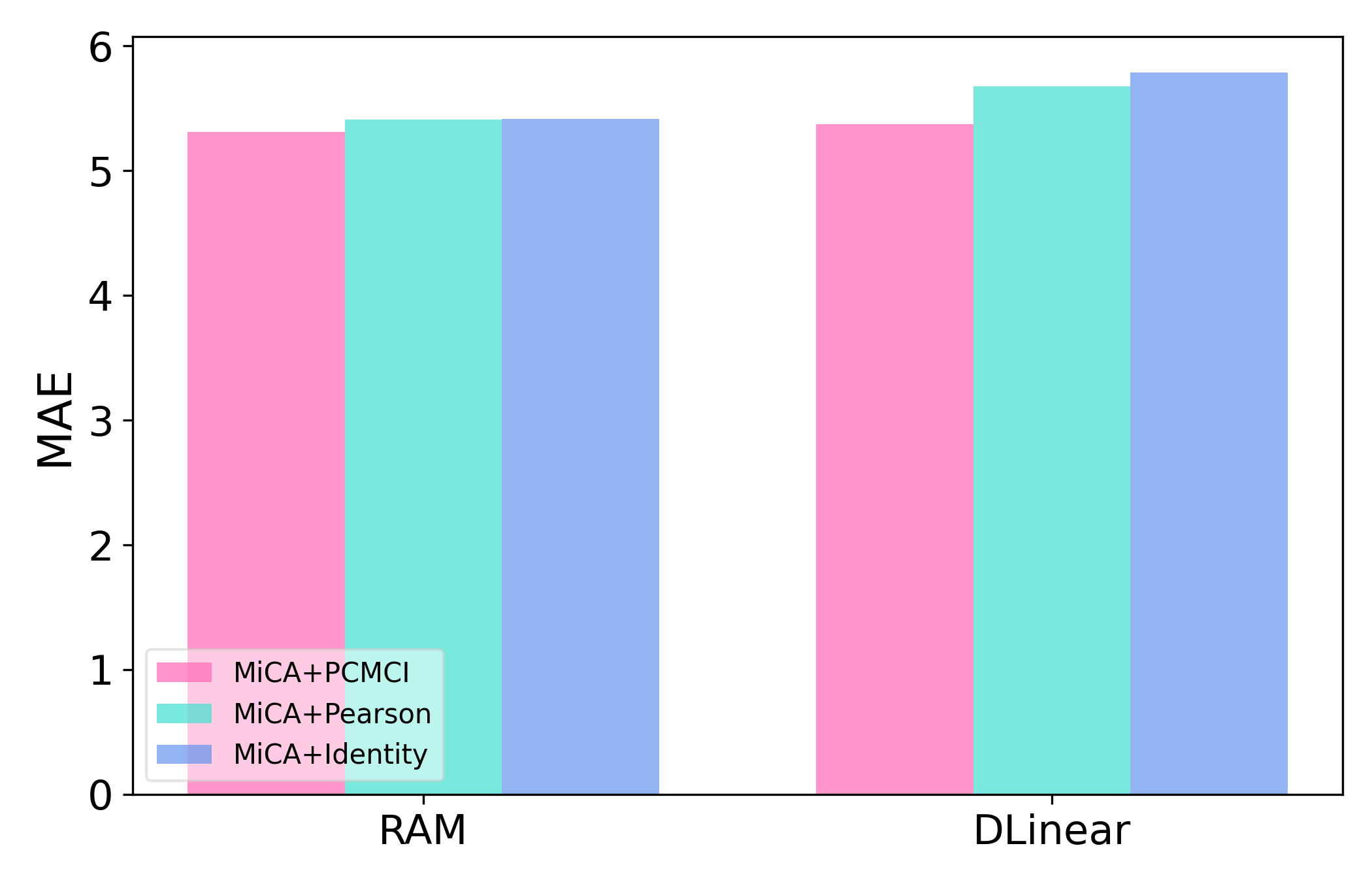}
        \caption{MAE}
        \label{fig:PCMCI_mae}
    \end{subfigure}
  
    \caption{Ablation analysis on causal discovery of the proposed model.}
    \label{fig:abl_PCMCI}
\end{figure}

\subsubsection{Causal Discovery}
Figure~\ref{fig:abl_PCMCI} evaluates the impact of different mobility priors used within MiCA by replacing the PCMCI-derived causal graph with either a Pearson correlation matrix or an identity matrix. For the RAM backbone, using Pearson correlations instead of PCMCI increases RMSE and MAE by 0.47\% and 1.88\%, respectively, while the identity prior leads to larger degradations of 0.51\% (RMSE) and 1.98\% (MAE). The effect is more pronounced for DLinear, where replacing PCMCI with Pearson correlations results in error increases of 4.42\% (RMSE) and 5.71\% (MAE), and using an identity prior further degrades performance by 6.14\% and 7.71\%. These results show that MiCA benefits not merely from adding spatial mixing, but from incorporating structured and directional mobility information. More informative priors yield larger gains, especially for highly lightweight backbones that lack intrinsic capacity to recover cross-regional dependencies from data alone.

\begin{table}[t]
\centering
\resizebox{0.3\textwidth}{!}{\begin{tabular}{@{}lrr@{}}
\toprule
Methods & FLOPs & Params \\ \midrule
DCRNN & 350.20M & 65.00B \\
STGCN & 110.83M & 103.92K \\
ColaGNN & 8.11M & \textbf{8.46K} \\ \midrule
RAM & 29.61M & 144.65K \\
RAM+MiCA & 48.01M & 482.47K \\
PatchTST & 86.69M & 407.82K \\
PatchTST+MiCA & 105.09M & 745.63K \\
DLinear & 22.57K & 406.00B \\
DLinear+MiCA & \textbf{1.54M} & 28.91K \\ \bottomrule
\end{tabular}}
\caption{Computational complexity comparison in terms of FLOPs and parameter count.}
\label{tab:abl_parameter}
\end{table}

\subsubsection{Model Complexity}
Table~\ref{tab:abl_parameter} reports the computational cost of MiCA in terms of FLOPs and parameter count. Although MiCA increases computation relative to the temporal encoder, the absolute overhead remains modest. For the RAM-pruned PatchTST backbone (denoted as RAM), MiCA adds 18.4M FLOPs and 138K parameters, resulting in 48.0M FLOPs and 482K parameters. This cost is still well below common spatio-temporal baselines such as STGCN (110.8M FLOPs) and DCRNN (350.2M FLOPs). Importantly, even after adding MiCA, RAM+MiCA remains substantially smaller than the original PatchTST backbone, with a 44.6\% reduction in parameter count. For PatchTST, MiCA increases FLOPs by 21.2\%, but the resulting model is still cheaper than graph-based baselines. For DLinear, the relative increase appears larger because the backbone is extremely small, yet the absolute cost remains negligible (1.55M FLOPs and 28.9K parameters). Notably, DLinear+MiCA achieves the lowest FLOPs among all evaluated models while maintaining competitive forecasting performance. Overall, these results show that MiCA provides a favorable trade-off between accuracy gains and computational overhead, enabling mobility-aware forecasting without resorting to heavy attention stacks or graph neural networks.

\section{Conclusion}
Accurate epidemic forecasting is challenging due to short, noisy surveillance data and the difficulty of reliably integrating indirect mobility signals. Existing mobility-aware models often rely on heavy relational architectures or abundant clean data, limiting their effectiveness in real-world public health settings. We proposed MiCA, a lightweight mobility-informed causal adapter that integrates mobility-derived structure into temporal forecasters through gated residual mixing. MiCA enables selective use of spatial dependencies while remaining robust under noisy and data-limited conditions. Experiments on four real-world epidemic datasets show that MiCA consistently improves lightweight backbones, often matching or outperforming parameter-heavy spatio-temporal baselines with substantially lower computational cost. Ablation and complexity analyses further confirm that these gains stem from causal mobility integration rather than increased model capacity. 

Overall, this work demonstrates that incorporating causal mobility priors as a soft structural prior can substantially enhance epidemic prediction without sacrificing efficiency. MiCA offers a practical framework for adapting modern benchmark-driven forecasting models to epidemiological settings, highlighting the broader potential of lightweight, causally informed approaches in high-stakes health applications.

\bibliographystyle{named}
\bibliography{COVID19_Mobility_Causal_Discovery}

\end{document}